# A Comparative study Between Fuzzy Clustering Algorithm and Hard Clustering Algorithm

Dibya Jyoti Bora[1]
Dr. Anil Kumar Gupta[2]
[1] Department Of Computer Science And Applications, Barkatullah University, Bhopal, India
[2] Department Of Computer Science And Applications, Barkatullah University, Bhopal, India

**Abstract:**

*Data clustering is an important area of data mining. This is an unsupervised study where data of similar types are put into one cluster while data of another types are put into different cluster. Fuzzy C means is a very important clustering technique based on fuzzy logic. Also we have some hard clustering techniques available like K-means among the popular ones. In this paper a comparative study is done between Fuzzy clustering algorithm and hard clustering algorithm.*

**Keywords:** *Clustering, FCM, K-Means, Matlab*

## 1. Introduction:

Data clustering is recognized as an important area of data mining [1]. This is the process of dividing data elements into different groups (known as clusters) in such a way that the elements within a group possess high similarity while they differ from the elements in a different group. Means to say that the whole clustering process should follow the following two properties: 1) High Intra cluster property and 2) Low inter cluster property.

Clustering can be classified as:

*Soft Clustering (Overlapping Clustering) & Hard Clustering (or Exclusive Clustering):*

In case of soft clustering techniques, fuzzy sets are used to cluster data, so that each point may belong to two or more clusters with different degrees of membership. In this case, data will be associated to an appropriate membership value. In many situations, fuzzy clustering is more natural than hard clustering.

Objects on the boundaries between several classes are not forced to fully belong to one of the classes, but rather are assigned membership degrees between 0 and 1 indicating their partial membership. On the contrary, in hard clustering techniques, data are grouped in an exclusive way, so that if a certain datum belongs to a definite cluster then it could not be included in another cluster. Fuzzy C Means (FCM) is a very popular soft clustering technique, and similarly K-means is an important hard clustering technique. In this paper, first of all, a detailed discussion on each of these two algorithms is presented. After that, a comparative study between them is done experimentally. On the basis of the result found, a conclusion is then drawn for the comparison.

## 2. Fuzzy C Means Clustering:

*Fuzzy C-means* (FCM) is a data clustering technique wherein each data point belongs to a cluster to some degree that is specified by a membership grade. This technique was originally introduced by Jim Bezdek in 1981 [4] as an improvement on earlier clustering methods [3]. It provides a method of how to group data points that populate some multidimensional space into a specific number of different clusters. The main advantage of fuzzy c – means clustering is that it allows gradual memberships of data points to clusters measured as degrees in [0,1]. This gives the flexibility to express that data points can belong to more than one cluster.

It is based on minimization of the following objective function:

$$J_m = \sum_{i=1}^{N}\sum_{j=1}^{C} u_{ij}^m \left\| x_i - c_j \right\|^2$$

, $1 \leq m < \infty$

where $m$ is any real number greater than 1, $u_{ij}$ is the degree of membership of $x_i$ in the cluster $j$, $x_i$ is the $i$th of d-dimensional measured data, $c_j$ is the d-dimension center of the cluster, and $\|*\|$ is any norm expressing the similarity between any measured data and the center. Fuzzy partitioning is carried out through an iterative optimization of the objective function shown above, with the update of membership $u_{ij}$ and the cluster centers $c_j$ by:

$$u_{ij} = \frac{1}{\sum_{k=1}^{C}\left(\frac{\|x_i - c_j\|}{\|x_i - c_k\|}\right)^{\frac{2}{m-1}}}$$

$$c_j = \frac{\sum_{i=1}^{N} u_{ij}^m \cdot x_i}{\sum_{i=1}^{N} u_{ij}^m}$$

,

This iteration will stop when $\max_{ij}\left\{\left|u_{ij}^{(k+1)} - u_{ij}^{(k)}\right|\right\} < \varepsilon$, where $\varepsilon$ is a termination criterion between 0 and 1, whereas $k$ are the iteration steps. This procedure converges to a local minimum or a saddle point of $J_m$.

The formal algorithm is :

1. *Initialize U=[$u_{ij}$] matrix, $U^{(0)}$*

2. *At k-step: calculate the centers vectors $C^{(k)}$=[$c_j$] with $U^{(k)}$*

$$c_j = \frac{\sum_{i=1}^{N} u_{ij}^m \cdot x_i}{\sum_{i=1}^{N} u_{ij}^m}$$

3. *Update $U^{(k)}$, $U^{(k+1)}$*

$$u_{ij} = \frac{1}{\sum_{k=1}^{C}\left(\frac{\|x_i - c_j\|}{\|x_i - c_k\|}\right)^{\frac{2}{m-1}}}$$

4. *If $\|U^{(k+1)} - U^{(k)}\| < \varepsilon$ then STOP; otherwise return to step 2.*

In FCM, data are bound to each cluster by means of a Membership Function, which represents the fuzzy behavior of this algorithm. To do that, we simply have to build an appropriate matrix named U whose factors are numbers between 0 and 1, and represent the degree of membership between data and centers of clusters.

## 3. K-Means Algorithm:

The K-Means [2] is one of the famous hard clustering algorithm [5][6][7]. It takes the input parameter $k$, the number of clusters, and partitions a set of $n$ objects into $k$ clusters so that the resulting intra-cluster similarity is high but the inter-cluster similarity is low. The main idea is to define k centroids, one for each cluster. These centroids should be placed in a cunning way because of different location causes different results. So, the better choice is to place them as much as possible far away from each other. The next step is to take each point belonging to a given data set and associate it to the nearest centroid. When no point is pending, the first step is completed and an early groupage is done. At this point we need to re-calculate k new centroids. After we have these k new centroids, a new binding has to be done between the same data set points and the nearest new centroid. A loop has been generated. As a result of this loop we may notice that the k centroids change their location step by step until no more changes are done. In other words centroids do not move any more. Finally, this algorithm aims at minimizing an *objective function*, in this case a squared error function. The objective function

$$J = \sum_{j=1}^{k} \sum_{i=1}^{n} \left\| x_i^{(j)} - c_j \right\|^2,$$

where $\left\| x_i^{(j)} - c_j \right\|^2$ is a chosen distance measure between a data point $x_i^{(j)}$ and the cluster centre $c_j$, is an indicator of the distance of the *n* data points from their respective cluster centers.

The Formal Algorithm [7] is:

1. *Select K points as initial centroids.*
2. *Repeat.*
3. *Form k clusters by assigning all points to the closest centroid.*
4. *Recompute the centroid of each cluster.*
5. *Until the centroids do not change.*

K-means algorithm is significantly sensitive to the initial randomly selected cluster centers. The algorithm can be run multiple times to reduce this effect. The K-Means is a simple algorithm that has been adapted to many problem domains and it is a good candidate to work for a randomly generated data points.

## 4. Experimental Results:

We chose Matlab for our experiments. The experiments done are mentioned below:

4.1 Matlab Implementation of K-means algorithm:

We have "kmeans" function to perform K-means clustering in Matlab [8]. The function kmeans performs K-Means clustering, using an iterative algorithm that assigns objects to clusters so that the sum of distances from each object to its cluster centroid, over all clusters, is a minimum. k means returns an n-by-1 vector IDX containing the cluster indices of each point. By default, kmeans uses squared Euclidean distances. When X is a vector, kmeans treats it as an *n*-by-1 data matrix, regardless of its orientation.

[IDX,C] = kmeans(X,k) returns the k cluster centroid locations in the k-by-p matrix C.

[IDX,C,sumd] = kmeans(X,k) returns the within-cluster sums of point-to-centroid distances in the 1-by-k vector sumd.

[IDX,C,sumd,D] = kmeans(X,k) returns distances from each point to every centroid in the n-by-k matrix D.

"iris " dataset [9][11] is chosen for the purpose of our experiment. The Iris flower data set or Fisher's Iris data set (some times also known as Anderson's Iris data) is a multivariate data set introduced by Sir Ronald Fisher (1936) as an example of discriminant analysis. The data set consists of 50 samples from each of three species of Iris (Iris setosa, Iris virginica and Iris versicolor). Four features were measured from each sample: the length and the width of the sepals and petals, in centimeters[9].

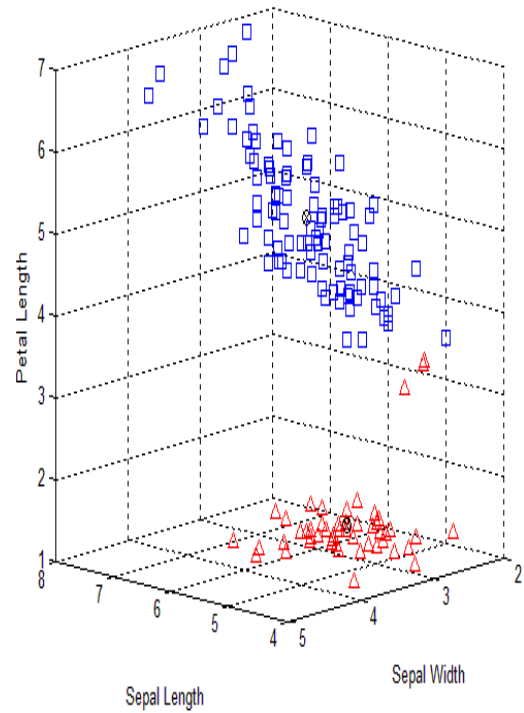

*Figure1: A 3-D plot of Iris data*

We specify 4 clusters and 5 replicates. Also 'display' parameter is used for printing out final sum of distances for each of the solutions. Replicate parameter is used to avoid local minima. The code for this purpose and with result obtained is as follows:

[cidx3,cmeans3,sumd3] = kmeans(meas,4,'replicates',5,'display','final');

Replicate 1, 6 iterations, total sum of distances = 71.7595.

Replicate 2, 8 iterations, total sum of distances = 57.2656.

Replicate 3, 4 iterations, total sum of distances = 57.2656.

Replicate 4, 16 iterations, total sum of distances = 71.4452.

Replicate 5, 17 iterations, total sum of distances = 57.2656.

Best total sum of distances = 57.2656

So, we found the best total sum of distances as 57.2285, and total elapsed time required is 1.4531 seconds.

4.2 Matlab Implementation of Fuzzy C Means:

In Matlab, Fuzzy C Means clustering can be performed with the function "fcm". This function can be described as follows[10]:

[center, U, obj_fcm] = fcm(data,cluster_ n)

The arguments of this function are:

1) *data* - lots of data to be clustering, each line describes a point in a multidimensional feature space;

2) *cluster_n* - number of clusters (more than one).

The function returns the following parameters:

1) *center* - the matrix of cluster centers, where each row contains the coordinates of the center of an individual cluster;

2) *U* - resulting matrix;

3) *obj_fcn* - the objective function value at each iteration

This means as input arguments, fcm takes a data set and a desired number of clusters. As output, it returns optimal clusters center, the resulting matrix U and the value of the objective function at each iteration. For our experiment, the same iris data set [9][11] is taken because, to compare the performance of K-means algorithm and Fuzzy C Means algorithm, the experiments should be performed on the same dataset. The data to be clustered is 4-dimensional data and represents sepal length, sepal width, petal length, and petal width. From each of the three groups (setosa, versicolor and virginica), two characteristics (for example, sepal length vs. sepal width) of the flowers are plotted in a 2-dimensional plot.

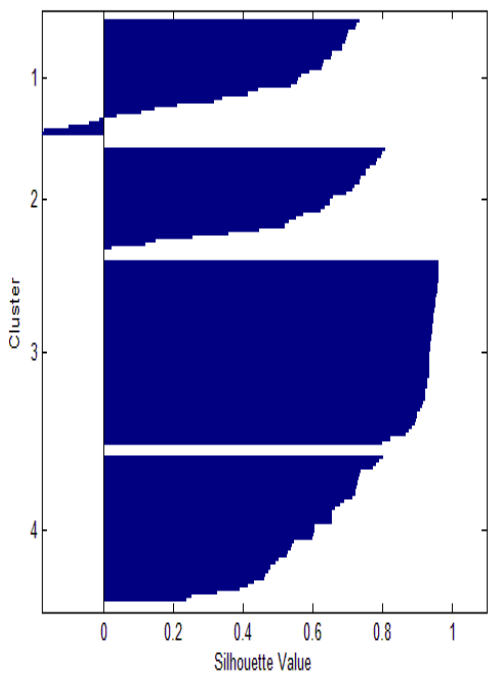

*Figure 2: Plotting of Four Clusters returned by kmeans function*

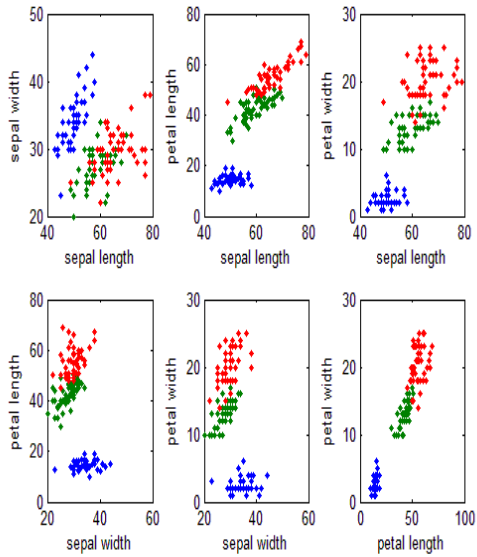

*Figure 3 : A Two Dimensional Plot Of The Iris Dataset*

We initialized the total number of clusters as 4 , maximum iteration no 100, exponent for U as 2.0 and minimum improvement as 1e-6. After initializing these values, we run the Fuzzy C Means. At maximum Iteration count = 28, the value of the obj. fcn obtained is 4168.707061. The total elapsed time taken is 4 seconds.

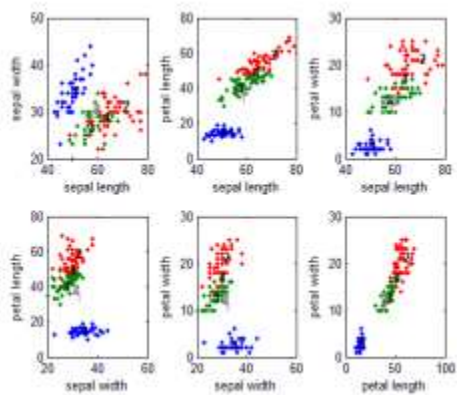

*Figure 4: Figure showing initial and final clusters*

## 5. Comparison of FCM and K-Means done on the basis of Experimental Results:

Comparison between FCM and K- Means algorithms is done on the basis of their respective computation times taken for the experiments and on the basis of their respective time complexities. First of all, it is merely visible from the observations that FCM algorithm is taking more time for computation than that of K-Means.

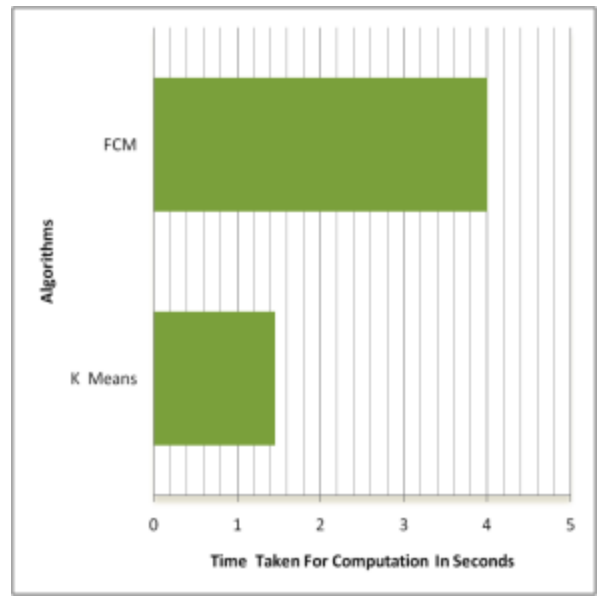

*Figure 5: FCM Vs K-Means in terms Elapsed Time*

The time complexity of K-Means algorithm is $O(ncdi)$ and time complexity of FCM is $O(ndc^2i)$[12][13][14].

Here n is the number of data points, c is the number of clusters, d is the dimension of the data and i is the number of iterations. Say n= 200, c = 1 to 4, d = 4, i= 28. Then, we have :

| Exp No. | No. Of Clusters | K-Means Complexity | FCM Complexity |
|---|---|---|---|
| 1 | 1 | 22400 | 22400 |
| 2 | 2 | 44800 | 89600 |
| 3 | 3 | 67200 | 201600 |
| 4 | 4 | 89600 | 358400 |

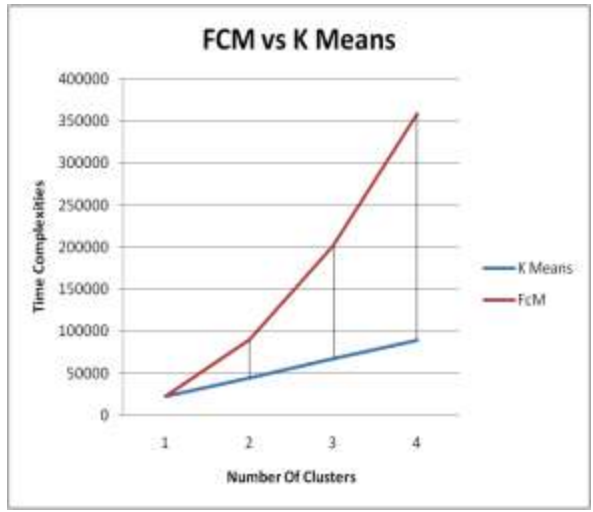

*Figure 6: FCM Vs K-Means in terms Time Complexities with Varying Number Of Clusters*

So, we have seen that, as soon as the number of cluster increases, the time complexity of FCM increases with a more rapid growth rate than that of K-Means algorithm. This draws the conclusion that K- Means algorithm is less complex than FCM.

## 6. Conclusion:

Choosing a particular clustering algorithm is solely dependent on the type of the data to be clustered and the purpose of the clustering applications. Hard clustering algorithm like K-Means algorithm is suitable for exclusive clustering task; on the other hand, fuzzy clustering algorithm like FCM is suitable for overlapping clustering task. In some situations, we cannot directly consider that data belongs to only one cluster. It may be possible that some data's properties contribute to more than one cluster. Like in case of document clustering, a particular document may be categorized into two different categories. For those purposes, we generally prefer membership value based clustering like FCM. In this paper, we have gone through a comparative research between Fuzzy clustering algorithm and Hard clustering algorithm.FCM is chosen on the behalf of Fuzzy clustering algorithm and K-Means algorithm is chosen on the behalf of Hard clustering algorithm. On the basis of experiments, we have found that the computational time of K-Means algorithm is less than that of FCM algorithm for the Iris dataset. So, this research work concludes the fact that the K-Mean's performance is better than FCM's performance in terms computational time. Since, fuzzy clustering algorithm includes much more fuzzy logic based calculations, so its computational time increases comparatively.